# Particle Swarm Optimization Framework for Low Power Testing of VLSI Circuits

Balwnder Singh[1], Sukhleen Bindra Narang[2] and Arun Khosla[3]

ACS Division-Centre for Development of Advanced Computing (CDAC), Mohali, India[1]
Electronics Technology Department, Guru Nanak Dev University, Amritsar, India[2]
ECE Department Dr. B .R. Ambedkar National Institute of Technology, Jalandhar, India[3]

`balwinder.cdacmohali@gmail.com`

## ABSTRACT

*Power dissipation in sequential circuits is due to increased toggling count of Circuit under Test, which depends upon test vectors applied. If successive test vectors sequences have more toggling nature then it is sure that toggling rate of flip flops is higher. Higher toggling for flip flops results more power dissipation. To overcome this problem, one method is to use GA to have test vectors of high fault coverage in short interval, followed by Hamming distance management on test patterns. This approach is time consuming and needs more efforts. Another method which is purposed in this paper is a PSO based Frame Work to optimize power dissipation. Here target is to set the entire test vector in a frame for time period 'T', so that the frame consists of all those vectors strings which not only provide high fault coverage but also arrange vectors in frame to produce minimum toggling.*

## KEYWORDS

*PSO (Particle Swarm Optimization), VLSI test, Low power, CUT (Circuit under Test), Frame work, Test Vectors, Fitness value, DUT (Design under Test).*

## 1. INTRODUCTION

In VLSI testing, in the initial years main concerned is the high fault coverage**.** But with the advance of technology focus is shifted towards the new target that is of having reduced test data volume, reduced test time and low power testing etc. The power which is dissipated during testing of a circuit is found of much higher value compared to the normal mode of operation. Main reason behind this problem is higher toggling count of the memory element like flip flops. So target in testing of sequential circuit is to have such sequences of test vectors which have high fault coverage, but also have an optimized toggling count. In testing easiest way to optimize the power is to reorder the test vectors of the test sequence, by doing this the test vectors are not modified, the test coverage is preserved.

PSO is the algorithm which is much similar to the genetic algorithm, but with lesser steps to be followed to reach the required destination. It does not include the steps like selection criteria in GA and crossover. PSO involves two parameters: One is '*p*best' and another is '*g*best'. '*p*best' stands for personal best that is to select the best position for the bits in the individual vector, and '*g*best' is to have vectors of best fitness function in the neighbor of the '*p*best'.

PSO optimizes the problems by following the population based search. The population is consists of random particles, which are randomly initialized and applied for solution of the problem. During simulation particles revise its own velocity and position based on the fitness

  

function of its own and entire population. In general PSO, particles move from one position to next best position, by updating itself on the basis of the past experience according to equation- 'a'& 'b'. In PSO individual chromosome mutate to move more near to the best solution for that particular individual, on the basis of the fitness value returned by the fitness function written to achieve a particular target. In PSO position and velocity of the moving swarm is updated continuously at a regular interval and in case of testing the position means the location of the bits in the test vector to have high fault coverage and velocity factor here in this paper is replaced by factor named toggling count, discussed in next sections.

v[] = v[] + c1 * rand() * (pbest[] - present[]) + c2 * rand() * (gbest[] - present[])  (a)
present[] = persent[] + v[] (b)

where, v[] is the particle velocity, persent[] is the current particle (solution). pbest[] and gbest[] are defined as stated before. rand () is a random number between (0,1). c1, c2 are learning factors. usually c1 = c2 = 2.[13]

In testing PSO particles population is in the binary numbers format which moves from one bit position to another to have such arrangement of bit patterns that will give maximum fault coverage for that particular collection of bit vectors, which is called '*p*best'. It is the same result which is got in GA, but the only difference is that not need to follow roulette wheel selection and crossover. Power is directly related to toggling factor according to formula 1.

$$p = \left(\frac{V_{dd}^2}{2 * clock\ period}\right) * \left(\sum_{for\ all\ gates\ g} \lfloor toggle(g) * C(g) \rfloor\right) - 1\ [6]$$

In Section 2 the previous work is presented. Under section 3 the idea of using PSO for optimized power for testing is given and includes the disadvantages of using PSO algorithm. Implementation of PSO for s27 and its results are given in section 4 followed by conclusion and Future scope.

## 2. PREVIOUS WORK

*K. Chakrapani et al.* presents the drawbacks of PSO algorithm if used for the purpose of power management during SoC teting. *Yanli Hou et al.* gives a new approach named as PSO based STPG (sequential Test Pattern Generator). In this paper the problem addressed is the optimization (PSO) and fault simulation for sequential circuits. Emphasize is given on three aspects: initialization, test sequence generation and test set compaction. *Dr. Karl O. Jones* in his paper discusses the comparison of genetic algorithm and PSO algorithm. *Dong Hwa Kim* uses the PSO to improve the local optimum solution of GA to have Euclidian data distance. *Wang jian et al.* presents a hybrid technique consisting of PSO and Ant algorithm to achieve higher fault coverage at faster rate. The hybrid algorithm based on particle swarm and Ant algorithm not only maintain the characteristic of simple and easy to realize of PSO algorithm, but also quicken the velocity of evolution and improve the precision of convergence under the condition of not adding running time evidently. *Krishna Kumar S et al.* presents the PSO for vector reordering to sort the problem of NP-complete by using Set of particles that correspond to states in an optimization problem, moving each individual in a numerical space looking for the optimal position. *Sandeep Singh Gill et al.* uses the hybrid technique of genetic algorithm and ant colony algorithm for the purpose of the VLSI circuits portioning, to sort out the NP hard problems.

## 3. PSO FRAMEWORK FOR POWER REDUCTION

During VLSI circuit testing preference is given to the techniques which are faster in respect of generating test vectors and fault propagation to the primary outputs. But at the same time there is a tradeoff between the power dissipation and the testing speed. Genetic algorithm is used for faster testing but it does not confirm about the low power dissipation, so hamming distance between the test vectors is reduced by rearranging the test sequences. Which cause the more time consumption for the testing of the particular DUT. But in PSO the above mentioned both techniques are merged into a one procedure. The steps followed in PSO are shown in the flow chart figure1

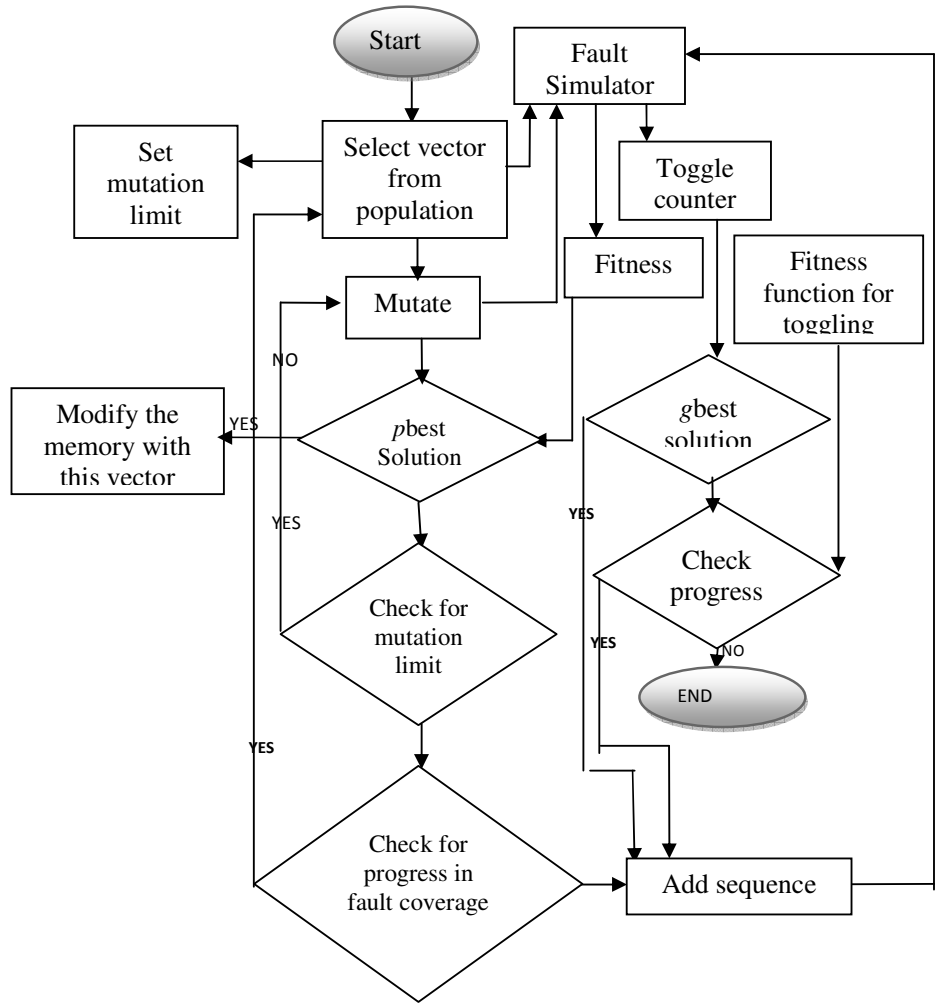

Figure 1: PSO flow for framing of test patterns

During implementation of PSO frame work, the common terms used in general PSO that are position and velocity are replaced with test vectors position and toggling count. Steps followed in PSO framing to have the test patterns in low power position with high fault coverage are:

**Step1:** Initialize the Random population.

**Step2:** Select a random vector from the population, and set the mutation limit.

**Step3:** For fault Propagation Simulate the CUT with the selected pattern.

**Step4:** check for '$p$best' based on the fitness value returned by the fault simulator.

15

**Step5:** Modify the memory according to fitness value and check for the mutation count, and mutate the vector again, and apply to fault simulator. Check for the fitness value again.

**Step6:** When no progress in the fitness values stopped, go for framing of the patterns.

**Step7:** Apply the frame to fault simulator and initialize the toggling counter. Compare the toggling value with the fitness value that is '$g$best'.

**Step8:** In case of improved '$g$best', keep the pattern in the frame format; otherwise remove the vector from the frame and select another vector form '$p$best' memory for framing [6].

Working of PSO framing logic for testing purpose is based on the mutation rate and the fitness function values returned from the fault simulator. But after some range mutation rate creates the problem and becomes the biggest disadvantage of this algorithm for testing purpose.

A PSO framing algorithm's pseudo code is given below.
```
Generate initial population
Cont_mutate='x';
Select the fault
Simulate ( For i=n;)
While (pbest)
      {Update the population};
While (mutate<='x')
         {do mutation}
              While(pbest)
      {Update the population};
If (progress)
      Got to random selection;
Else break;
Add sequence in list
Simulate (For m= sequence list ;)
Initialize Togel_count='p';
While( gbest)
      {keep sequence list}
Check progress;
If ('yes')
Take new sequence list;
Else
End;
```

Mapping from individual vector into the form of test sequence generation as illustrated in figure2, each character in population is applied to a primary input. And number of individuals in frame or in population varies as the number of the primary inputs varies from circuit to circuit.

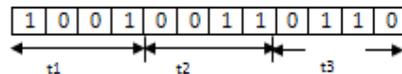
Figure2: Test sequence generation

If such a time frame encounters in which no faults are detected, then in search space many faults have to be dropped, which causes the less fault coverage.

An accurate fitness function is the key factor in case of having high quality test set frame. The second factor which affects the most is the mutation limit. If the limit is too small then there always exists the probability of better solution, but if mutation range approaches to higher value then the optimum solution then it starts diverting from the target or remains unaffected. So time is consumed more, many of the particles are wasting computational effort in seeking to move in the same direction towards the local optimum already discovered, whereas better results may be obtained if various particles explore other possible search directions. That means having same set of vectors again at continuous time interval.

When PSO is applied to a sequential circuit then first condition is to check the fitness of the applied vector on the basis the number of flop change their state. Pseudo code for this fitness function is given below.

```
N= number of vectors in population
Select a vector form the population;
Function flipflop_state( )
For (i=initialize the population;<=N;i++)
{where i is the initial vector and N is total number vectors in the population;
flip_flop( );
Retrun(state);}
End for; end function;
```

When all the flip flops are set then the fitness of the all vectors on the basis of the number of fault propagated to primary output is calculated and arranged in the decreasing order after mutation process is applied, and fitness is again verified. This gives the new population with elevated fitness. Alongside a counter is set which calculate the number of non- contributing vectors whose fitness value is below specified or required range.

When the number of noncontributing vectors generated exceeds the limit, and then it means there is no further progress in '$p$best' then PSO proceeds with test sequence generation which is obtained with help of '$g$best'. Here the fitness functions and the procedure followed is presented in pseudo code as illustrated above. Here a frame is modeled by adding a sequence as shown in figure 2, if the frame causes less toggling, calculated from the value returned by the toggling counter then the vector in the sequence is kept otherwise detached from sequence and new vector length is added. As at the end of the process, based on the '$g$best' value the final frame is the of less toggling and high fault coverage format. That frame is applied at different clock cycles that's is t1, t2, and t3, as shown in an example in figure2

## 4. Experimental Results

S27 consists of 10 gates and 3 D-flip-flops. Here to initial the flip flop's initial state these are assumed as a direct primary input lines as presented in the figure 3. To generate random population of test vector command line based Turbo Tester tool is used. And random population which has obtained is shown in table 1. Number of faults forced are 32. Random population includes 9 test patterns for this benchmark circuit.

After applying PSO farming the resulted six test patterns detect all 32 faults shown in table 2. From the '$g$best' fitness function of the PSO these vectors are arranged in a sequence or frame consisting reduced toggling value shown in table3. Fault coverage between the PSO based test patterns and random test vectors are plotted in figure4.

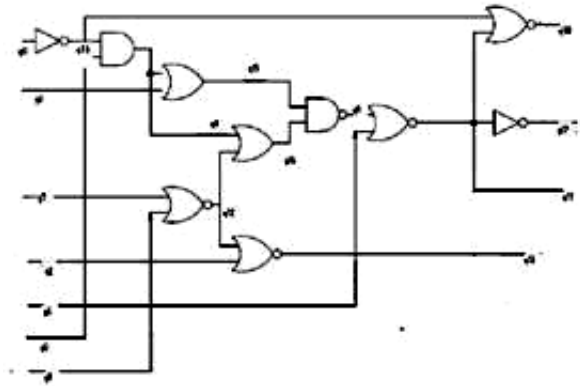

Figure.3 S27 benchmark circuits

Table 3 PSO framing with least toggling for s27

| 1101010 | 1001000 | 0000100 | 0110110 | 0010011 | 0010000 |
|---------|---------|---------|---------|---------|---------|
| T6 | T5 | T4 | T3 | T2 | T1 |

Table 1. Random vector list

| S.No. | Random test vectors | Fault coverage |
|-------|---------------------|----------------|
| 1 | 1110010 | 21.875 |
| 2 | 1001101 | 31.25 |
| 3 | 0100111 | 40.625 |
| 4 | 0010000 | 50.0 |
| 5 | 0010010 | 68.75 |
| 6 | 0101010 | 75.0 |
| 7 | 0001000 | 90.625 |
| 8 | 1011010 | 93.75 |
| 9 | 1101000 | 100.0 |

Table 2. List of vectors after PSO framing given by *p*best

| S.No. | Random test vectors | Fault coverage |
|-------|---------------------|----------------|
| 1 | 1101010 | 34.375 |
| 2 | 0110110 | 62.5 |
| 3 | 1001000 | 81.25 |
| 4 | 0000100 | 90.625 |
| 5 | 0010011 | 93.75 |
| 6 | 0010000 | 100.0 |

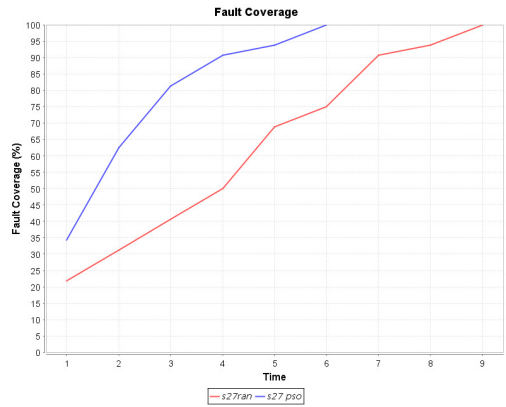

Figure 4. Fault coverage for S27 with and without PSO

Based on the sequence generated after applying PSO on the vectors of maximum fault coverage power dissipation is calculated by using equation 1 and power dissipated is 0.0975 *pico*watts, which half of 0.18 *pico*watt.

## 5. Conclusion:

A Technique based on the Particle Swarm Optimization Framework is utilized for achieving low toggling rate of flip-flops in the Circuit Under Test. '*P*best' factor in PSO gives the test vectors having higher fault coverage from the list of random test vector collection. A code is written in "C" for acquiring '*g*best' of that higher fault coverage test vectors and arranged into a sequence or a frame of minimum toggling, to attain minimum power during the testing period.